%% file: lifeplanner.tex
\pdfoutput=1

\documentclass[11pt]{article}

\usepackage{EMNLP2023}

\usepackage{times}
\usepackage{latexsym}

\usepackage[T1]{fontenc}

\usepackage[utf8]{inputenc}

\usepackage{microtype}

\usepackage{inconsolata}
\usepackage{balance}
\usepackage{url}
\usepackage{times}
\usepackage{latexsym}
\usepackage{xspace}
\usepackage{booktabs}

\usepackage{colortbl}
\usepackage[shortlabels]{enumitem}
\setlist[itemize]{leftmargin=*}
\setlist[enumerate]{leftmargin=*}
\usepackage[T1]{fontenc}

\usepackage[utf8]{inputenc}

\usepackage{microtype}
\usepackage{listings}

\usepackage{amsmath, amsfonts, amssymb, amsthm, mathtools}

\usepackage{graphicx} 

\newcommand{\sname}{\ensuremath{\mathsf{LLMPlan}}\xspace}
\newcommand{\og}{\ensuremath{\mathsf{SymPlan}}\xspace}
\newcommand{\llmog}{\ensuremath{\mathsf{SymPlan+}}\xspace}

\definecolor{mygreen}{RGB}{34, 139, 34}
\definecolor{myred}{RGB}{209, 0, 86}
\definecolor{myblue}{RGB}{110, 245, 227}

\definecolor{nav}{RGB}{0,0,128}

\title{Human-Centered Planning}

\author{Yuliang Li \and Nitin Kamra \and Ruta Desai \and Alon Halevy \\
\{yuliangli, nitinkamra, rutadesai, ayh\}@meta.com \\
Reality Labs Research, Meta}

\begin{document}

\maketitle

\begin{abstract}
LLMs have recently made impressive inroads on tasks whose output is structured, such as coding, robotic planning and querying databases. The vision of creating AI-powered personal assistants also involves creating structured outputs, such as a plan for one's day, or for an overseas trip. Here, since the plan is executed by a human, the output doesn't have to satisfy strict syntactic constraints. A useful assistant should also be able to incorporate vague constraints specified by the user in natural language. This makes LLMs an attractive option for planning.

We consider the problem of planning one's day. We develop an LLM-based planner (\sname) extended with the ability to self-reflect on its output and a symbolic planner (\og) with the ability to translate text constraints into a symbolic representation. Despite no formal specification of constraints, we find that \sname\ performs explicit constraint satisfaction akin to the traditional symbolic planners on average ($2\%$ performance difference), while retaining the reasoning of implicit requirements. Consequently, LLM-based planners outperform their symbolic counterparts in user satisfaction (70.5\% vs.\ 40.4\%) during interactive evaluation with 40 users.

\end{abstract}

\section{Introduction} \label{sec:intro}

\begin{figure*}
    \centering
    \includegraphics[width=0.98\textwidth]{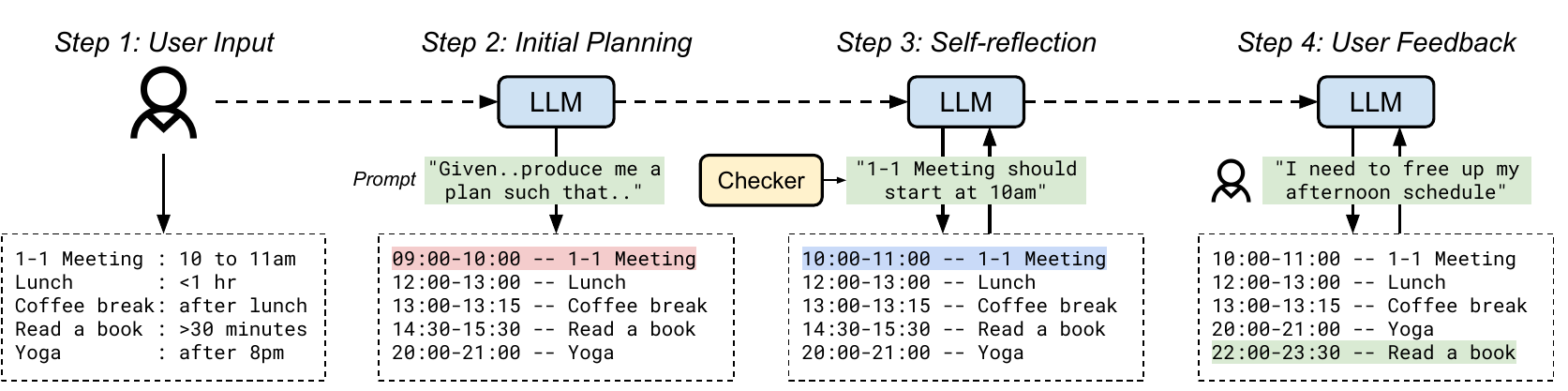}
    \caption{The \sname\ planner takes as input the events to be scheduled with their associated constraints. After extracting the plan from the LLM's response, the system self-reflects by checking for constraint violations and prompting the LLM to fix them. A user may interact with \sname\ in several iterations to refine their plan. In doing so, \sname\ is able to incorporate constraints that may be vague. We show that the accuracy of \sname\ is on par with a symbolic planner (\og) which is unable to incorporate vague constraints.}
    \label{fig:architecture}
\vspace{-3mm}
\end{figure*}

Inspired by the success of 
large language models (LLMs)  on many NLP  tasks, several research efforts have explored  leveraging LLMs for tasks whose outputs are structured, such as coding~\cite{chen2021evaluating}, robotics planning~\cite{DBLP:conf/icml/HuangAPM22}, and querying external data sources~\cite{schick2023toolformer,thoppilan2022lamda}. These tasks are more complicated because the outputs need to be syntactically well-structured (e.g., code, a plan for a robot, a SQL query) and they require the LLM to exhibit more logical reasoning, which is one of its known weaknesses.

Applying LLMs to structured tasks is an opportunity to reexamine the basic assumptions regarding such tasks.  The hallmark of structured tasks is that the algorithms designed to solve them provide correctness guarantees  on their output, but at the cost of having to completely specify the problem setting in the input.  For example, in planning problems, one must formally describe the possible states of the world, the actions that take us from one state to another, including the preconditions and postconditions of every action, and finally, we need to describe the  desirable goal states~\cite{silver2022pddl}.  However, in many real-world situations, especially when AI is applied to more human-centered tasks, it's impractical to completely describe the full richness of possible states of the world and the constraints on them.  At the same time, the outputs of the plan are carried out by humans, not machines, which means that they don't necessarily need to satisfy rigid syntactic constraints. Put together, the above observations suggest that there are new opportunities in applying LLMs to structured tasks.

Motivated by these opportunities and the vision of creating personal AI assistants, we consider the task of Day Planning. A plan for one's day needs to include the items on the user's todo list and calendar, in addition to things they do habitually (e.g., exercise, walk the dog). Items are typically associated with constraints on their duration, start and end time, or ordering w.r.t.\ other items (e.g., coffee {\em after} lunch). Some of these constraints may be vaguely or implicitly specified or omitted completely (e.g.,  need frequent breaks in the afternoon, go to the supermarket). 
Specifying all these considerations in a constraint language, as is needed by traditional symbolic planners, is hard to imagine. This is further complicated by the fact that these constraints are extremely personal. Instead, a system should just be able to understand such constraints and preferences specified in natural language. The same observations hold for other AI-assistant tasks, such as planning a trip or preparing a dinner party.

This paper explores the promise of using LLMs for day planning by comparing it with traditional symbolic planning. We begin with two planners,  an LLM-based planner (\sname) that is based on prompting an LLM and a symbolic planner (\og) that requires explicit constraints but is guaranteed to return a correct answer. We then explore a few hybrid variants. We extend \sname\ with the ability to self-reflect on its output and fix parts of the plan that contradict the input constraints. We extend \og\  with the ability to translate text constraints into a symbolic representation and to elicit commonsense constraints from an LLM. We compare these planners on a synthetic dataset that is generated from a few real-world planning examples and show that \sname\ with self reflection comes within 2\% accuracy of \og. However, unlike \og, \sname\ is able to also reason with implicit constraints. 

We then evaluate \sname and \og in an interactive setting with 40 users, where users refine their plans via multiple iterations with the planner. We find that $47\%$ of user-specified  constraints were vague and open-ended. Consequently, \sname outperforms \og in user satisfaction (70.5\% vs.\ 40.4\%), owing to its ability to handle such vagueness.  In conclusion, our experiments suggest that LLM-based planning offers a promising technique for personal AI assistants.

\input{related}

\input{method}

\section{Experiments} \label{sec:experiments}

We evaluate the different planners using both synthetic and real data. In summary, our results show that by leveraging LLMs and self-reflection, \sname can understand well-specified constraints and generate correct plans similar to a symbolic planner.
Meanwhile, \sname performs significantly better at filling in constraints based on commonsense knowledge when they are not explicitly specified in the input. Finally, our user study shows that \sname provides promising results for interacting with users to refine their constraints.


\subsection{Datasets}



\noindent
\textbf{Synthetic dataset.} Building realistic datasets 
for day planning is challenging. 
To do so, we first conducted a survey of 4 users collecting information about
(1) what events/activities they usually plan and (2) what constraints
they would like to specify on those events. 
To mimic realistic personas, we asked ChatGPT to provide other examples of  todos, habits, and pre-scheduled events that people care about, by prompting it with the initial set provided by these four individuals. 
Lastly, we uniformly sample from each set and ensemble the subsets into 100 synthetic
personas (See Appendix for more details).

\smallskip
\noindent
\textbf{Real-user dataset.} To obtain real user planning needs, 
we hosted a user-study.
Through the study, we collected 40 user profiles 
which we curated to form the real benchmark
(see Section \ref{sec:userstudy} for more details).

Table \ref{tab:benchmark} shows statistics of the 2 datasets
and Table \ref{tab:benchmark_examples} shows example events
and constraints. To the best of our knowledge, these
 are the first available benchmarks for day planning.


\begin{table}[!ht]
\small
\caption{\small Statistics of the two datasets. 
Note that the synthetic dataset has $\sim$2x the number of events and constraints
per user (the numbers in parentheses) compared to the real-user dataset.}
\label{tab:benchmark}
\centering
\begin{tabular}{cccc} \toprule
    & \#users & \#events & \#constraints \\ \midrule
Synthetic & 100     & 1,035 (10.35)    & 2,128 (21.28)   \\
Real-user & 40& 223 (5.575)     & 384 (9.6) \\ \bottomrule 
\end{tabular}
\vspace{-3mm}
\end{table}

\setlength{\tabcolsep}{2.5pt}
\begin{table}[!ht]
\caption{\small Example events and constraints from our datasets. 
Note that the real-user dataset may contain events that do not belong to the 3 pre-defined
categories.
}\label{tab:benchmark_examples}
\small
\begin{tabular}{ccc}\toprule
Event   & Example    & Example Constraints    \\ \midrule
ToDo    & Plan a vacation  & \textgreater{}1 hour, \textless{}2 hours \\
Habit   & grocery shopping & after meeting, before 17 \\
Pre-scheduled & Project Update   & starts at 10, ends at 11     \\
Real-user    & Play Xbox  & after dinner \\ \bottomrule 
\end{tabular}
\end{table}



\subsection{Evaluation Metrics}

Due to the complex and subjective nature of the task,
there is no single metric for evaluating the
quality of a plan. Instead, we consider a 
 set of metrics that provide a more comprehensive view of the behavior
of the planner.

\smallskip
\noindent
\textbf{Correctness metrics. } We first focus on a set
of metrics measuring the correctness of the plan, i.e., 
whether the planner understands the input constraints
and correctly follows them. For these metrics,
we expect a symbolic planner to achieve (close to) 
optimal performance. Ideally, the LLM-based planner should perform close to the symbolic one. 

\begin{itemize}[itemsep=1pt,topsep=2pt]
\item \textbf{Event Coverage (CO)}: percentage of input events included
in the plan,
\item \textbf{Non-Overlap (NO)}: percentage of planned events that don't overlap with others,  and
\item Percentage of \textbf{Duration constraints (DC), 
Order Constraints (OC),
Start/End constraints (SEC)} that are correctly satisfied.
\end{itemize}

\smallskip
\noindent
\textbf{Commonsense violations. }
We evaluate how well the planner understands commonsense
human needs. For this purpose, we curated a set of commonsense constraints
on event duration and start/end time, e.g., breakfast is 
usually in the morning (ends before 11am) and 
takes less than 1 hour. A movie night event is in the evening 
(i.e., after 5pm) and takes at least 90 minutes.
For each generated plan, we measure the percentage of
violations of the \textbf{duration} and \textbf{start/end} 
commonsense constraints. 

Note that these constraints 
are not hard constraints that the planners must satisfy.
For example, the planner may decide to schedule a language class
in the evening which is uncommon but may still be acceptable.
Achieving a low commonsense violation score
indicates that the planner is capable of capturing human behavioral
patterns, even when they are not explicitly specified in
the input. For this set of metrics, we expect the LLM-based
or the hybrid approach to outperform the symbolic planners that do not take event semantics into account.

\setlength{\tabcolsep}{4.5pt}
\begin{table*}[!ht]
\caption{Plan quality metrics for our synthetic dataset. }
\label{tab:results_synthetic}
\centering
\resizebox{.98\textwidth}{!}{
\begin{tabular}{c|cccccc|ccc} \toprule
  & \multicolumn{6}{c}{Correctness ($\uparrow$)}    & \multicolumn{3}{c}{Commonsense ($\downarrow$)} \\
Methods                         & CO    & NO     & DC    & OC    & SEC    & AVG   & duration & start\_end & AVG   \\ \midrule
GPT-3 (curie)                   & 23.91 & \textbf{100.00} & 40.00 & 84.72 & 14.29  & 52.58 & 
28.00 & 29.45 & 28.73  \\
GPT-3 (davinci)                 & 81.16 & 99.17  & 66.86 & 82.79 & 33.33  & 72.66 & 16.50    & 24.06     & 20.28 \\
GPT-3.5 (SFT, text-davinci-002) & 88.54 & 99.80  & 93.00 & 96.59 & 69.75  & 89.54 & 13.22    & 17.76     & 15.49 \\
GPT-3.5 (RL, text-davinci-003)  & 98.95 & 98.25  & 92.77 & 96.76 & 73.00  & 91.94 & 17.82    & 16.46     & 17.14 \\
GPT-3.5 + self-reflect          & 99.83 & 97.80  & 98.29 & 98.38 & 79.19  & 94.70 & 16.81    & 18.06     & 17.43 \\
ChatGPT                         & 88.35 & 98.68  & 95.05 & 94.60 & 82.98  & 91.93 & 11.19    & \textbf{15.06}     & 13.12 \\
ChatGPT + self-reflect          & 94.29 & 97.48  & 99.21 & 97.38 & 97.88  & 97.25 & 10.35    & 15.73     & 13.04 \\
GPT-4                           & 88.96 & 98.68  & 94.81 & 94.94 & 85.03  & 92.48 & 10.52    & 15.71     & 13.11 \\
GPT-4 + self-reflect            & 93.16 & 97.38  & \textbf{99.47} & 97.08 & 98.38  & 97.09 & \textbf{10.21}    & 15.38     & \textbf{12.80} \\ \midrule
\og                             & 99.72 & \textbf{100.00} & 95.90 & \textbf{99.68} & \textbf{100.00} & \textbf{99.06} & 24.36    & 34.42     & 29.39 \\
\llmog                          & \textbf{100.00}	& 98.22	& 98.56	& 97.66	& 99.51	& 98.79	& 25.88	& 12.33	& 19.11 \\ \bottomrule
\end{tabular}}
\end{table*}

\setlength{\tabcolsep}{4.5pt}
\begin{table*}[!ht]
\caption{Plan quality metrics for our dataset on real users.}
\label{tab:results_real}
\centering
\resizebox{.98\textwidth}{!}{
\begin{tabular}{c|cccccc|ccc} \toprule
  & \multicolumn{6}{c}{Correctness ($\uparrow$)}    & \multicolumn{3}{c}{Commonsense ($\downarrow$)} \\
Methods & CO    & NO    & DC    & OC    & SEC    & AVG   & duration  & start\_end  & AVG   \\ \midrule
GPT-3 (curie)                   & 21.58 & 100.00 & 78.57 & 67.92 & 33.33  & 60.28 & \textbf{5.67}     & 57.45     & 31.56 \\
GPT-3 (davinci)                 & 78.31 & 94.03  & 61.54 & 68.25 & 42.55  & 68.94 & 9.09     & 31.25     & 20.17 \\
GPT-3.5 (SFT, text-davinci-002) & 88.99 & 98.96  & 96.36 & 90.85 & 78.43  & 90.72 & 11.71    & 24.32     & 18.02 \\
GPT-3.5 (RL, text-davinci-003)  & 97.10 & 99.58  & 93.55 & 93.55 & 74.07  & 91.57 & 15.25    & 24.58     & 19.92 \\
GPT-3.5 + self-reflect          & 98.45 & 99.17  & \textbf{97.87} & 95.17 & 84.62  & 95.06 & 14.55    & 24.55     & 19.55 \\
ChatGPT                         & 82.86 & 98.20  & 94.64 & 96.83 & 95.12  & 93.53 & 9.71     & 17.48     & 13.59 \\
ChatGPT + self-reflect          & 88.88 & 96.22  & 97.67 & 97.67 & 93.75  & 94.84 & 11.70    & 17.02     & 14.36 \\
GPT-4                           & 82.65 & 97.43  & 94.44 & 97.58 & 95.12  & 93.44 & 8.82     & 17.65     & \textbf{13.24} \\
GPT-4 + self-reflect            & 89.19 & 98.04  & 97.67 & \textbf{98.40} & 94.12  & 95.49 & 10.42    & 17.71     & 14.06 \\ \midrule
\og                             & 98.92 & 98.90 & 95.24 & 97.26 & \textbf{100.00} & \textbf{98.06} & 13.33 & 35.83 & 24.58 \\
\llmog                          & \textbf{99.27} & \textbf{100.00} & 93.65 & 97.18 & 96.23 & 97.27 & 15.70 & \textbf{15.70} & 15.70 \\ \bottomrule
\end{tabular}}
\vspace{-2mm}
\end{table*}
\subsection{Implementation and Baselines}

We implemented \sname using Langchain\footnote{\url{https://langchain.readthedocs.io/en/latest/}}
and OpenAI's LLM APIs\footnote{\url{https://openai.com/api/}}. 
For the LLM-based planner, we evaluate various LLMs
to understand what techniques enable the planning 
capabilities:
\begin{itemize}[itemsep=1pt,topsep=2pt]
\item \textbf{GPT-3:} Transformer-based LLM pre-trained on 
web and general text data~\cite{DBLP:conf/nips/BrownMRSKDNSSAA20}.
We evaluate the base model \textsf{curie} of 6.7B parameters
and \textsf{davinci} model of 175B parameters.
\item \textbf{InstructGPT / GPT-3.5
\cite{DBLP:conf/nips/Ouyang0JAWMZASR22}: }
These models are instruction-tuned to enable better following of human instructions. 
We test the \textsf{text-davinci-002} model
based on supervised fine-tuning and
\textsf{text-davinci-003} 
based on reinforcement learning, both with 175B parameters.
\item \textbf{ChatGPT~\cite{chatgpt}:} 
The well-known chat version of GPT-3.5 optimized for dialog. We use the 
\textsf{gpt-3.5-turbo} checkpoint in our
experiment.
\item \textbf{GPT-4~\cite{DBLP:journals/corr/abs-2303-08774}:} 
The most recent OpenAI LLM known for its significantly improved 
capabilities on problem solving and handling complex 
instructions versus GPT-3.5 (but whose size is unknown). 


\end{itemize}

\noindent
\textbf{Self-reflection: } We also evaluate the self-reflection technique on
the top 3 best performing LLMs 
for planning, i.e., 
GPT-3.5 (\textsf{text-davinci-003}),
ChatGPT, and GPT-4.

\noindent \textbf{Symbolic planning: } 
\begin{itemize}[itemsep=1pt,topsep=2pt]
\item \textbf{\og:} Symbolic planning baseline described in section~\ref{sec:symbolic-planner}.
\item \textbf{\llmog:} Combines \og with commonsense constraints inferred using GPT-3.5 (\textsf{text-davinci-003}) as described in section~\ref{sec:symbolic-planner}.
\end{itemize}


\begin{figure*}[!ht]
    \centering
    \includegraphics[width=0.98\textwidth]{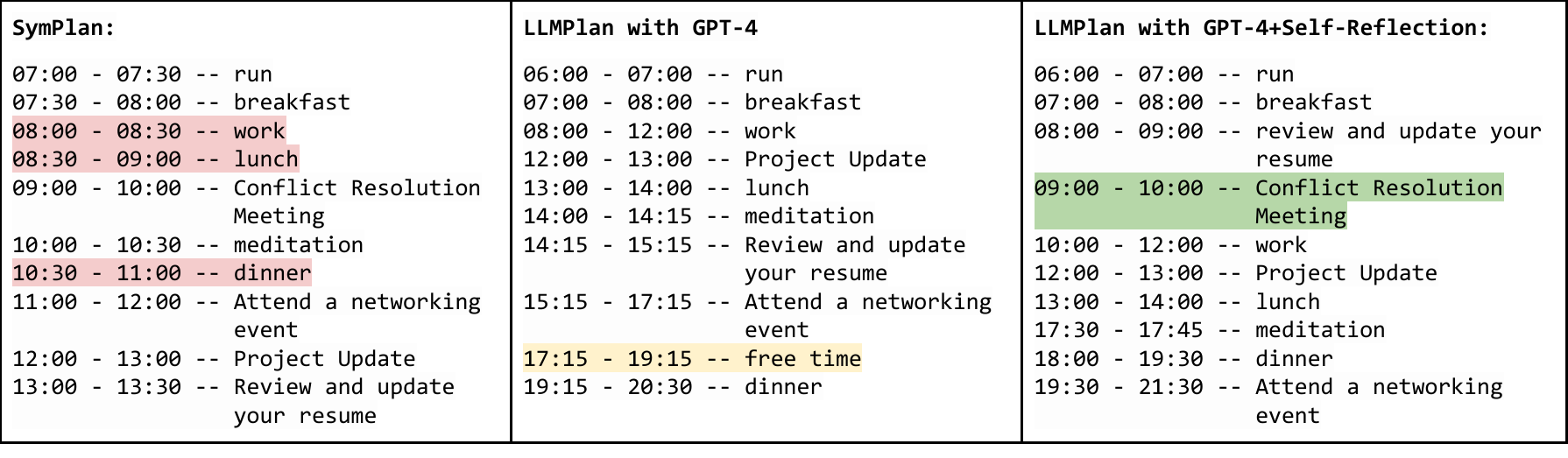}
    \caption{\small Output of 3 planners for the persona in Figure
    \ref{fig:example_input}. While \og satisfies all the input 
    constraints, it violates the time/duration commonsense
    for work and meals (in \textcolor{red}{red}). 
    LLM (GPT-4) generates a natural and mostly correct plan
    with 1 input event missing.
    Self-reflection correctly adds back the missing event 
    (in \textcolor{green}{green}).
    Note that LLM may generate novel events not from the input 
    such as ``free time'' (in \textcolor{yellow}{yellow}).}
    \label{fig:example}
\vspace{-3mm}
\end{figure*}

\begin{figure}[!ht]
    \centering
    \includegraphics[width=0.48\textwidth]{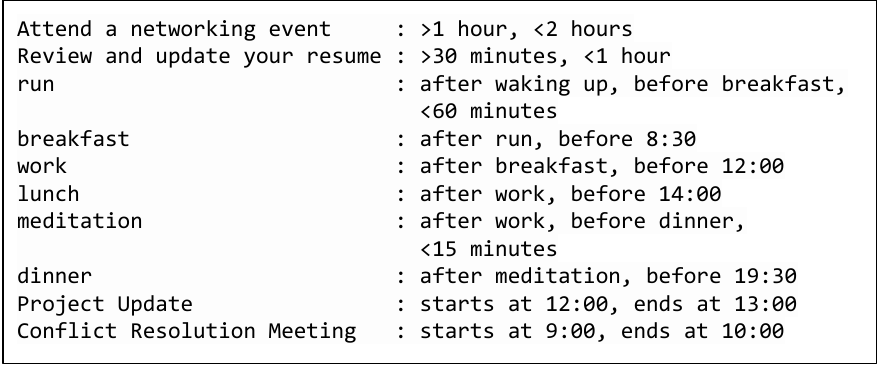}
    \caption{\small An example persona from our synthetic dataset with various events and their constraints. }
    \label{fig:example_input}
\vspace{-3mm}
\end{figure}

\subsection{Results}

\textbf{For LLM-based planning, model size and instruction tuning are critical for planning.}  
As seen in Tables \ref{tab:results_synthetic} and \ref{tab:results_real}, the GPT-3.5 models outperform GPT-3 on both
correctness and commonsense understanding by large margins,
e.g., by over 40\% average correctness score
or 5\% to 10\% commonsense scores on both benchmarks. Overall, the recent large and instruction-tuned LLMs outperform their counterparts. 

\smallskip
\noindent \textbf{Self-reflection 
effectively improves the correctness metrics.} It improves GPT-3.5 on
4/5 correctness metrics by an additional 2.87\% on average 
in the real benchmark and improves ChatGPT on
3/5 correctness metrics by an additional 2.27\% on average 
in the synthetic benchmark.

\smallskip
\noindent \textbf{Symbolic planners excel at correctness but suffer at common-sense reasoning.}
\og achieves an average correctness score of 99.06\%
on the synthetic set but shows a performance gap of 10-15$\%$ on common-sense reasoning from best-performing LLM.

\smallskip
\noindent \textbf{Hybrid approaches achieve a reasonable trade-off between correctness and commonsense reasoning.}
\llmog improved the commonsense capabilities of \og by up to 10$\%$, without 
affecting its correctness scores. Likewise, GPT-4 with self-reflection, 
outperforms \og significantly on commonsense scores (>10\%)
while closing the performance gap on correctness
(over 95\% on both benchmarks). 
Overall, GPT-4 with self-reflection is the best LLM-based method in terms of both correctness and commonsense scores. 



\begin{table*}[!ht]
\small
\caption{\small Breakdown of all 113 user requirements into 3 categories: well-defined,
open-ended, and complex.} \label{tab:breakdown}
\begin{tabular}{ccc} \toprule
Category     & Count & Examples                                                                                     \\ \midrule
Well-defined & 59    & "add a phone call at 10:00", "shorten my lunch to 30 minutes", "End the last event before 20:00"                                \\
Open-ended    & 40    & "schedule meal breaks", "there will be a rain around 10am, what should i prepare?"           \\
Complex      & 14    & "I need to break up the demos into two meetings", "I want at least 5 minutes between events" \\ \bottomrule
\end{tabular}
\vspace{-3mm}
\end{table*}

\begin{table}[!ht]
\small
\centering
\caption{\small Overview of user feedback.} \label{tab:feedback}
\resizebox{.48\textwidth}{!}{
\begin{tabular}{ccc|cc|cc} \toprule
         & \multicolumn{2}{c|}{\sname} & \multicolumn{2}{c|}{\llmog} & \multicolumn{2}{c}{Both} \\ 
Feedback & Pos.          & Neg.          & Pos.          & Neg.         & Pos.         & Neg.        \\ \midrule
Count    & 31           & 13           & 17           & 25          & 11          & 9          \\ 
\%	& \textbf{70.5\%} & 29.5\% & 40.5\% & 59.5\% & 55\% & 45\% \\
\bottomrule
\end{tabular}}
\end{table}

\begin{table}[!ht]
\small
\centering
\caption{\small The numbers of requirements per category and their corresponding user feedback are shown. \sname performs well on
the open-ended requirements, while the complex ones are the difficult for both planners.}
\label{tab:messages}
\begin{tabular}{ccccccc} \toprule
             & \multicolumn{2}{c}{\sname} & \multicolumn{2}{c}{\llmog} & \multicolumn{2}{c}{Overall} \\ 
           Category  & Pos.          & Neg.          & Pos.         & Neg.        & Pos.         & Neg.         \\ \midrule
Well-defined & 36            & 23            & 24           & 35          & 60           & 58           \\
Open-ended      & 29            & 11            & 18           & 22          & 47           & 33           \\
Complex    & 7             & 7             & 5            & 9           & 12           & 16 \\ \bottomrule
\end{tabular}
\vspace{-2mm}
\end{table}

\subsection{Error analysis}

We  manually inspected the plans for  errors 
made by the planners. For \sname, the most common errors were (1) to hallucinate events that were not in the input (e.g., add ``free time'' at various points in the day) and (2) to slightly rename events (e.g., rename ``take marketing class'' to  
``Take a marketing class'') or combine multiple events into one. The renaming of events is not a strict error, and even the combination of events often made sense and satisfied the timing constraints. 
\og, on the other hand, sometimes returned plans that were not very natural, such as putting a meal at a completely wrong time (when appropriate constraints were not given). 
 We show in Figure \ref{fig:example_input}
and Figure \ref{fig:example}  examples of typical behaviors of the planners.

\subsection{Evaluation with users}\label{sec:userstudy}

Our experiments  so far focus on evaluating plans objectively, without any user feedback. To better understand the utility of \sname and \og, 
 we conduct a user study where users can refine their plan by iteratively providing feedback to the planner.  In doing so, we also obtain insights into the kinds of constraints and events that users specify to a day planner. 
For example, after observing an initial plan, the user may want to re-plan 
with requests in natural language 
such as ``can you move event X to the afternoon?''.
Sometimes the requests can be more complicated, such as
``I want frequent breaks in the afternoon''.

To support this experiment, we enhanced both planners with additional features
for incorporating user feedback.  For \sname this merely involved putting it behind a chatbot interface.  The chatbot receives messages from the user and
appends them as additional input to the LLM to update the plan (see Figure \ref{fig:gptprompt} for the prompting detail).

For  \llmog, we developed an extractor, that given a user statement such as  ``lunch should start at noon and be at least 30 minutes'', would try to identify constraints that can be added to the planner's input, for example,

\begin{center}\small
\texttt{``lunch: starts at 12:00, >30 minutes''} 
\end{center}
to update the original input to \llmog. Note that the interactive capability
of this approach is limited by the expressive power of the current
constraint language and the extractor.

\smallskip
\noindent
\textbf{Setup. } Our participants included designers, engineers, and researchers. 
 Each session starts 
with an introduction to the planner UI (see Figure \ref{fig:ui} in the Appendix).
Then the user enters  a set of events that they want to plan
and then runs both planners. After observing the initial plans,
the user interacts with the planner via commands to the chatbot.
Finally, they provide feedback on whether they like/dislike
the final plan.
Each session lasted for $\sim$15 minutes.

\smallskip
\noindent
\textbf{Results. } During the user study, we collected 40 valid sessions 
(i.e., the user uses both planners and gives at least 1 feedback).
Note that the 40 personas from the real benchmark are taken
from these 40 valid sessions (see Table \ref{tab:benchmark}).
Table \ref{tab:feedback} shows the summary of the collected user feedback.
Overall, \sname received more positive feedback versus \llmog (70.5\% vs. 40.5\%).

An interesting discovery from the user study is the types of requirements that users provide to the planner.  
We collected 113 user requirements in total. We organize them into 3 heuristically derived categories (see Table \ref{tab:breakdown} for some examples and stats):
\begin{itemize}[itemsep=1pt,topsep=2pt]
\item \textbf{Well-defined:} (52.2\%) these are human instructions with clearly defined
semantics of action, 
such as add/change/remove events or constraints. 

\item \textbf{Complex:} (12.4\%) these are messages with clear semantics but requiring
the planner to execute complex steps, which are beyond
the capabilities of current natural language understanding (NLU) approaches.
\item \textbf{Open-ended:} (35.4\%) these are messages that are ambiguous requiring
additional human context to execute correctly. NLU alone will not be
sufficient.
\end{itemize}
In Table \ref{tab:messages}, we measure how well the planners can correctly
handle each category of messages. 
We found that \sname significantly outperforms \og on handling the
open-ended messages (72.5\% vs. 45\% positive rate), while
the complex messages remain the most difficult type with no more than 50\%
positive rate for both planners.


\input{conclusion}

\section{Limitations}


\noindent
\textbf{Evaluation. }
Our current evaluation with users is small scale due to the high cost for collecting real user input
and/or feedback. While scaling up the evaluation on synthetic data is possible, it still requires human effort on verifying/annotating the commonsense constraints.

\smallskip
\noindent
\textbf{LLM usage. } The current implementations
of both \sname and \llmog are based on OpenAI's
LLM's through API access. We observed that
these APIs can be slow/unstable and may
incur a high cost for large-scale experiments.
We are currently exploring the option
of running the experiments on locally hosted
LLMs such as LlaMA~\cite{thoppilan2022lamda}.

\smallskip
\noindent
\textbf{Prompting vs. fine-tuning.} 
An explored promising direction to explore is 
fine-tuning a LLM for the planning task, 
probably via  instruction tuning. This can
potentially reduce the inference cost using a 
smaller base LLM (e.g., LlaMA-7b) and significantly
speed up inference time.

\section{Ethics}
\smallskip
\noindent
\textbf{Usage of personal data.} 
Day planning is an inherently personal task requiring private data. We erred on the side of caution and predominantly leveraged synthetic data without any personally identifiable information (PII) but grounded in reality where possible. Our real-user dataset was carefully collected with user consent and without any PII under an approved protocol. 
Future real-world day planning applications would certainly require additional work to ensure that personal data are secure and being meaningfully and responsibly used.  

\newpage

\bibliographystyle{acl_natbib}

\bibliography{lifeplanner}

\appendix

\section{Prompts for LLM-based planner}

\begin{figure*}
    \centering
    \includegraphics[width=0.90\textwidth]{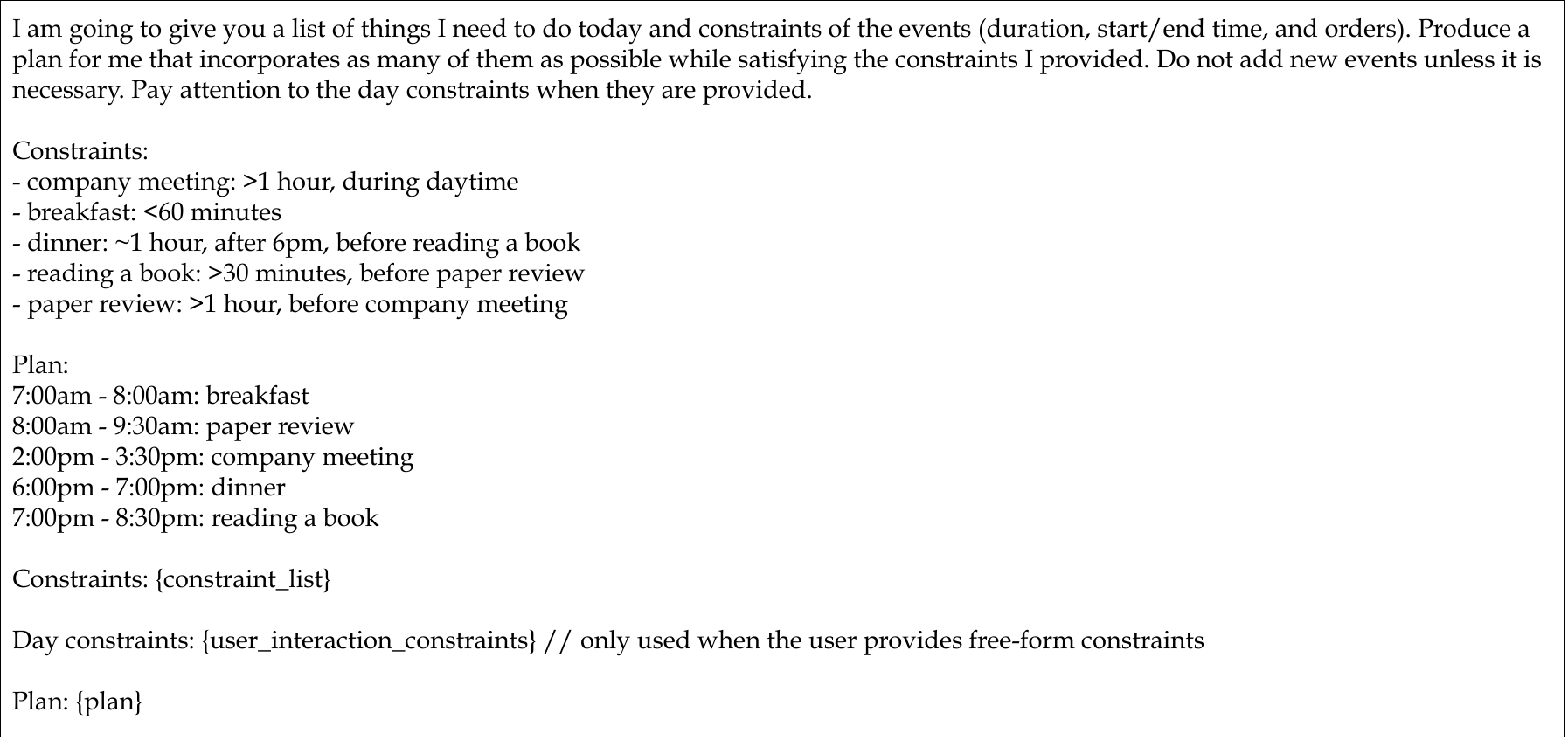}
    \caption{Prompt for day plan generation given constraints.}
    \label{fig:gptprompt}
\end{figure*}

Figure \ref{fig:gptprompt} shows the input prompt that \sname
used to generate the initial day plan give the events and their
constraints as input.

\begin{figure*}
    \centering
    \includegraphics[width=0.90\textwidth]{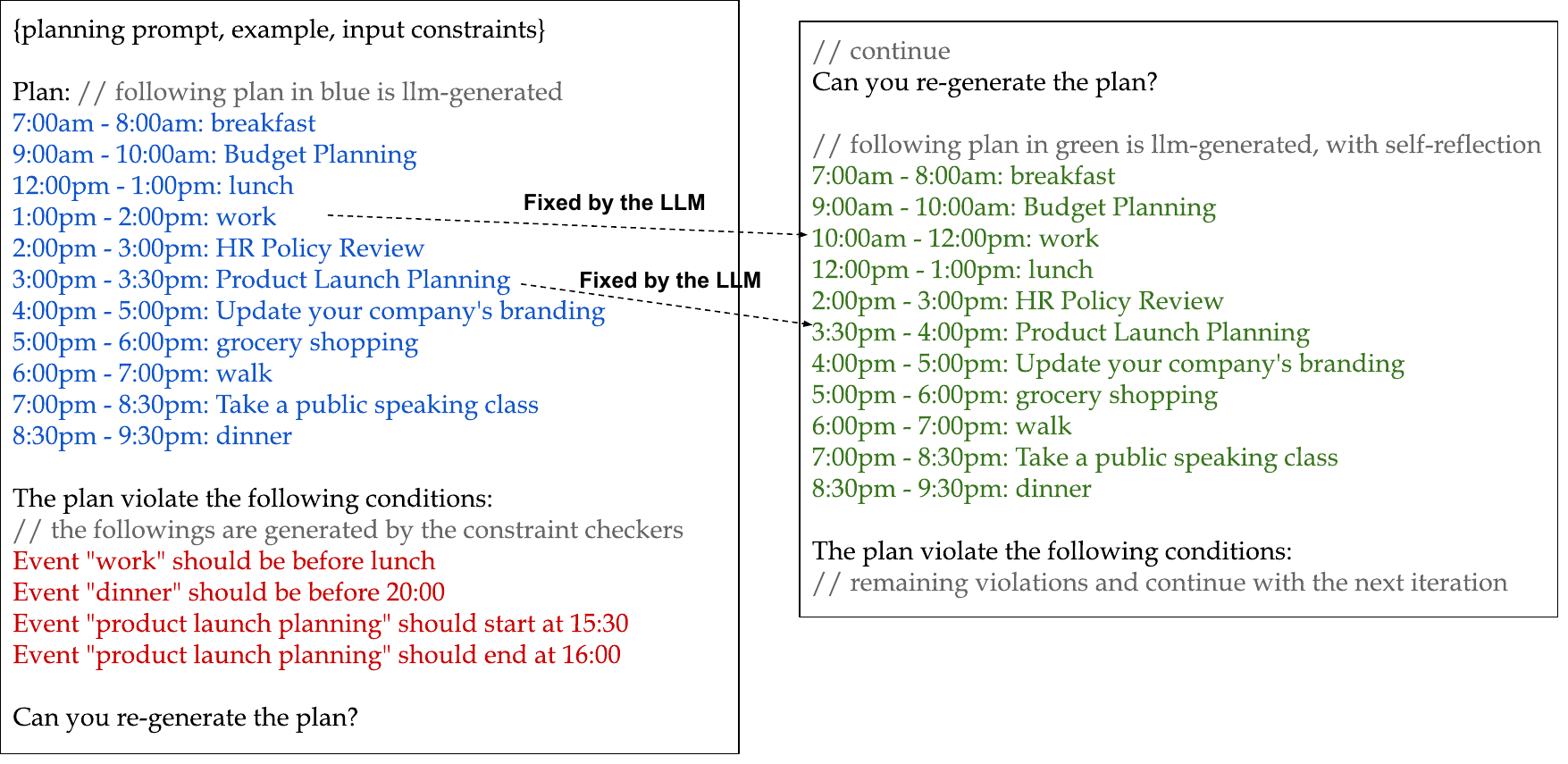}
    \caption{An example of \sname with self-reflection.}
    \label{fig:self_reflection}
\end{figure*}

Figure \ref{fig:self_reflection} shows an example of using the
self-reflection prompt to automatically correct constraint violations
for an LLM-generated plan. In this example, the original generated
plan (in \textcolor{blue}{blue})
violates 4 constraints (in \textcolor{red}{red}). The self-reflection
prompt sends the these 4 violations back to the LLM as feedback
and asks it to re-generate. The new plan generates a new plan
(in \textcolor{green}{green})
with 3 of the violations (1 for ``work'' and 2 for ``lunch'') fixed.
This process continues with the remaining violations as input
until there is no violation or it reaches the maximum number of iterations.

\section{Details for the synthetic dataset}


We build each persona from the synthetic dataset by 
generating random samples of todos, habits, and pre-scheduled events.

\smallskip
\noindent
\textbf{ToDos: } We first generate a candidate set of todos by
using seed events from 4 users. We feed the seed events to ChatGPT\footnote{https://chat.openai.com/chat} to
generate (1) a candidate list of 78 todo events and (2) the min/max
duration of the event to form the duration constraints. 
For each persona, we select 2 events from the candidate list uniformly
at random.

\begin{figure}
    \centering
    \includegraphics[width=0.48\textwidth]{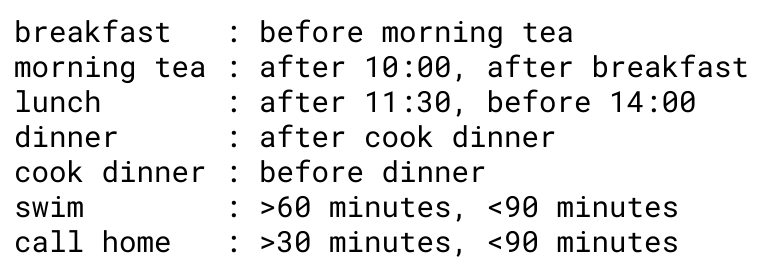}
    \caption{An example seed habit persona.}
    \label{fig:example_habits}
\end{figure}

\smallskip
\noindent
\textbf{Habits: } From the same 4 users, we collected samples
of their habits. See Figure \ref{fig:example_habits} for an example.
From the 4 habit personas, we use ChatGPT to generate 50 more instances
by mimicking the pattern. We randomly select 1 instance for each generating
each persona.

\smallskip
\noindent
\textbf{Pre-scheduled: } We first generate 200 titles of the pre-scheduled
events using a similar method as the one for ToDos. We then randomly generate
their start/end time uniformly from the range of [9:00, 17:00]. Each
event can be either 1-hour or 30-minute long with a probability of 1/2.
Each generated persona contains 3 randomly generated pre-scheduled events.

\section{Details of \og}
\label{sec:app_og_plan}

In this section, we describe the temporal scheduling based symbolic planner: \og in more detail. Given the set of events $E$ and constraints $L_e$ for each event $e \in E$, \og computes the start and end times: $\mathsf{start}_e$ and $\mathsf{end}_e$ of events by solving the following optimization problem:
\begin{align*}
\min & \{ \max_{e \in E} \mathsf{end}_e \} \\
\text{s.t.} \\
& l_{e,i} \text{  holds, } & \forall l_{e,i} \in L_e, \hspace{8pt} \forall e \in E.
\end{align*}
The above translates to minimizing the end time of the final task (i.e. $\max_{e \in E} \mathsf{end}_e$) while enforcing the specified temporal constraints ($l_{e,i}$).

Since it may not be possible to enforce all the constraints for the tasks together, \og takes a two-phase approach:
\begin{enumerate}
    \item Constraint checking and relaxation,
    \item Backtracking search.
\end{enumerate}

\textbf{Constraint checking and relaxation}: The first phase encodes all the temporal constraints into the distance graph representation of a simple temporal network (STN)~\cite{dechter1991temporal}. Figure~\ref{fig:stp} in the main paper shows an example with start and end nodes for two events ($\mathsf{Lunch}$ and $\mathsf{Jog}$) in an STN along with the datum node which represents $t=0$ (12:00am) in our case. The edges in the graph represent temporal constraints. For details about STNs and distance graphs, we refer the interested reader to \citet{dechter1991temporal}.

This distance graph representation is useful since running a one-to-all shortest path algorithm from the $\mathsf{datum}$ to all other nodes gives us an upper-bound on time for scheduling each node. Similarly, by running an all-to-one shortest path algorithm from all nodes to the $\mathsf{datum}$ gives a lower-bound on time for scheduling each node. We do the former via the Bellman-ford algorithm~\cite{cormen2022introduction} and the latter by reversing all the edges in the graph and then another run of the Bellman-ford algorithm. If we get valid lower and upper bounds from this step: $\mathsf{start}_{e,lb}, \mathsf{end}_{e,ub}$ for each event $e$, we proceed to the second phase of backtracking search. However, if the Bellman-ford algorithm detects negative cost cycles in the graph, it indicates that the constraints are infeasible.

For detecting and relaxing edges which cause infeasibilities, we adopt a simple approach relying on iterative addition of edges in the distance graph~\cite{oei2009resolving}. We create a copy of the distance graph with just the nodes. We then add the edges, one at a time while detecting negative cost cycles, if any, using an iterative version of the Bellman-ford algorithm similar to the one proposed by \citet{cesta1996gaining} for efficiency. Any edge that is found to induce infeasibility is relaxed up to $K$ times by changing its weight by a pre-specified amount $\Delta t$\footnote{We use $K=16$ and $\Delta t=15$ minutes in our experiments.}. If this still does not fix the induced infeasibility, the constraining edge is dropped altogether.

\textbf{Backtracking search}: Having obtained valid time bounds for scheduling each node and having asserted feasibility with the specified temporal constraints, we next generate a plan using backtracking search. Note that no events are allowed to overlap in time and this is implicitly enforced during search.

The search state maintains a list of events allocated to the user along with their assigned start and end times, ordered by their start times. At each level of the search, we consider all unallocated events so far and identify the ones which can be feasibly allocated next. An event $e$ could be infeasible if: (a) it participates in a relative temporal constraint with a predecessor event and its predecessor has not been allocated yet, or (b) if the last allocated event ends after the time upper-bound to start $e$: $\mathsf{start}_{e,ub}$. Next we compute the \textit{slack} of all identified feasible events. We define \textit{slack} of an event $e$ as the amount of time remaining before it becomes infeasible to allocate, i.e., by the difference between $\mathsf{start}_{e,ub}$ for the event $e$ and the end time of the last allocated event.

Considering the feasible events in ascending order of their \textit{slack}, the next feasible event is allocated to the user's plan while assigning to it its earliest possible start and end times. Then the search state is updated and all the upper and lower bounds for each node in the distance graph are tightened using our iterative version of the Bellman-ford algorithm. The search method is then recursively called with the newly updated allocated event list. When there are no more feasible events left, we record a solution (or a partial solution if some events were left unallocated altogether). The last allocated event is then de-allocated, the nodes' upper and lower bounds restored to their values before the allocation and the search backtracks. Upon completion of the search or on expiry of a $max\_search\_time$ cutoff\footnote{We use $max\_search\_time=1$ second in experiments.}, we return the best full solution (or best partial solution, if no full solution was found) found till then.

\section{Details of \llmog}
\label{sec:app_gptog_plan}

\llmog currently uses GPT-3.5 for the LLM prompts to extract commonsense constraints. The LLM prompt consists of three few-shot examples followed by a query for the event of interest. We present the LLM prompt below, where \{event\} is replaced by the actual event being queried for:
\vspace{4pt}
\par \noindent\fbox{
    \parbox{\linewidth}{\small
        Can you suggest typical start and end time constraints during a day for the following event:\\
        dinner\\

        Constraints:\\
        dinner: starts after 18:00, ends before 22:00\\

        Can you suggest typical start and end time constraints during a day for the following event:\\
        wake up\\

        Constraints:\\
        wake up: starts after 7:00\\
        
        Can you suggest typical start and end time constraints during a day for the following event:\\
        company meeting\\

        Constraints:\\
        company meeting: starts after 10:00, ends before 17:00\\

        Can you suggest typical start and end time constraints during a day for the following event:\\
        \{event\}\\

        Constraints:
    }
}
\vspace{4pt}

The response produced by the LLM for the example where \{event\}=$\mathsf{write\ my\ paper}$ is shown below:
\par \noindent\fbox{
    \parbox{\linewidth}{\small
        write my paper: starts after 13:00, ends before 17:00
    }
}

\section{Planner UI}

\begin{figure*}
    \centering
    \includegraphics[width=0.97\textwidth]{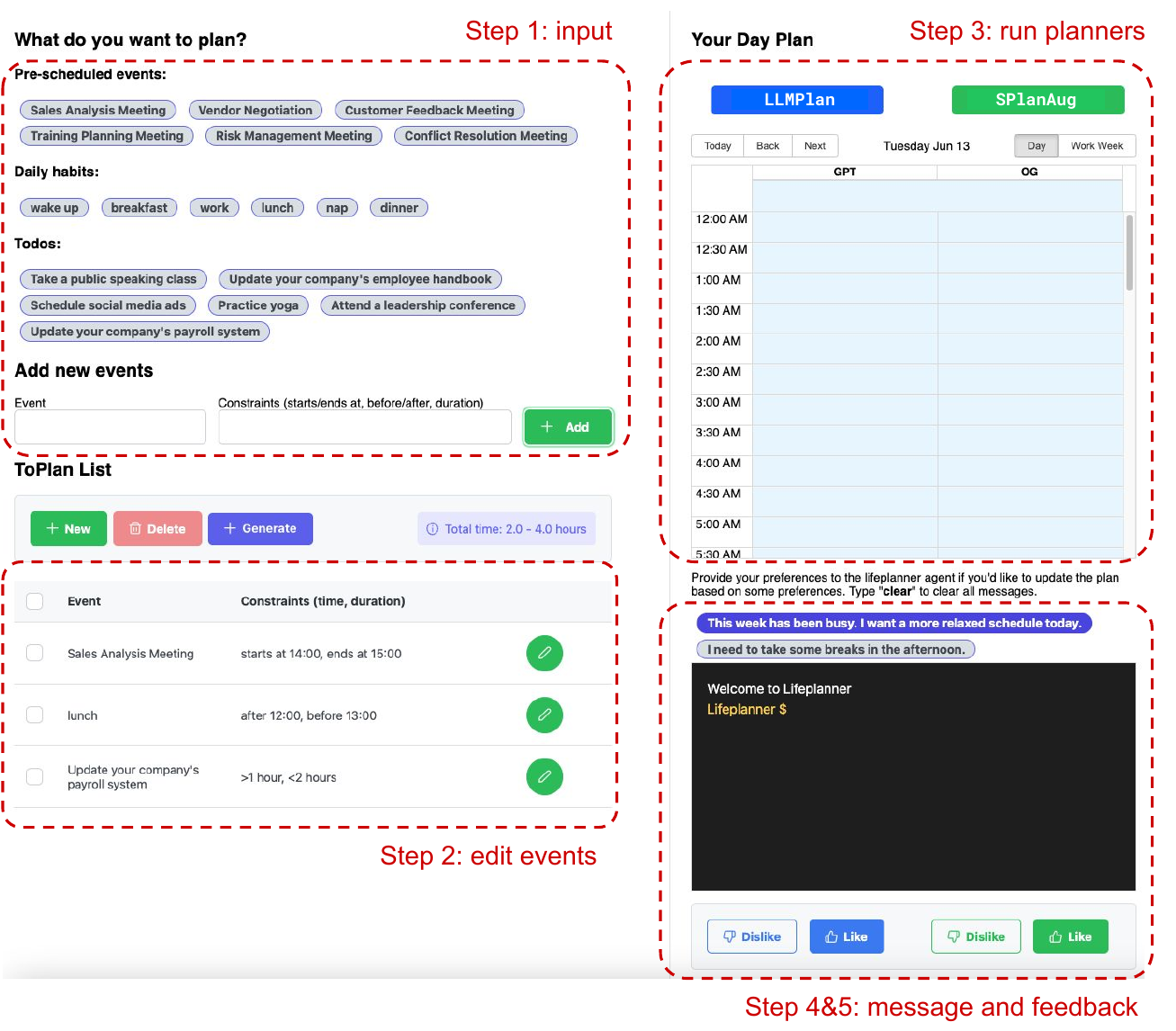}
    \caption{The planner UI with steps involved in the user-study sessions.}
    \label{fig:ui}
\end{figure*}

Figure \ref{fig:ui} shows the UI design for the user
study in Section \ref{sec:experiments}. The UI has 5 main
components: (1) an input panel, (2) an event editing area,
(3) a calendar view for the two planners (\sname and \llmog),
(4) a chatbot interface for iterating over the plans generated by the planners, and
(5) a widget for collecting binary feedback for both planners.

\end{document}

%% file: related.tex
\section{Related work} \label{sec:related}
\noindent
\textbf{LLMs for planning.}
Recently, there has been a lot of interest in leveraging LLMs for planning. Researchers have explored prompting and in-context learning~\cite{huang2022language, singh2022progprompt, wang2023describe}, use of feedback~\cite{huang2022inner, wang2023voyager, wang2023describe}, as well as use of reflection, self-debugging, and self-critique~\cite{shinn2023reflexion, chen2023teaching, wang2023voyager, kim2023language} for plan generation. The majority of these focus on planning for embodied agents in closed-loop simulators or real-world settings. While we also leverage in-context learning and self-reflection, our goal is to generate \emph{schedules} for \emph{humans} -- where the generation must account for the duration of actions and their start and end times. Unlike embodied AI settings, which predominantly require grounded reasoning of affordances and executability, our problem necessitates reasoning of ambiguous temporal constraints. Personal scheduling and planning for humans have also been investigated in Human-Computer Interaction (HCI) domains for assistive applications~\cite{bilbily2021space, myers2007intelligent}. However, to the best of our knowledge, our work is the first to consider it in the context of LLM-based reasoning.

Researchers have also explored the use of LLMs for generalized planning i.e., generating plans for PDDL domains~\cite{silver2022pddl, valmeekam2023planning, pallagani2022plansformer}. Rather than performing PDDL planning with LLMs, 
translation of natural language to PDDL followed by symbolic planning has also been investigated~\cite{liu2023llm+, collins2022structured, xie2023translating}. Our hybrid \llmog approach is similar in spirit, where we extract constraints and common-sense requirements from natural language for symbolic planning downstream, albeit without PDDL.

\noindent
\textbf{Classical scheduling.} Day planning has traditionally been approached as a temporal scheduling problem that models temporal constraints within and between events~\cite{saksena1993temporal}. Most major algorithms to solve this problem rely on either compiling these constraints into a simple temporal network (STN) and using solution approaches based on shortest-path computations~\cite{dechter1991temporal} or expressing the constraints as linear inequalities and using linear programming based approaches~\cite{kreter2016models} or constraint satisfaction algorithms~\cite{alexiadis2009defining}. Our \og algorithm is similar in spirit to the STN model adopted by \citet{dechter1991temporal}, but also augments it by including mechanisms to detect feasibility~\cite{cesta1996gaining} and constraint relaxation~\cite{oei2009resolving}.



%% file: method.tex
\section{Planning a day} \label{sec:gpt}

We begin by defining the Day Planning problem. 

\subsection*{Day plans}
Given a set of activities/tasks that the user wants to accomplish during the day,  
the goal of the planner is to generate an agenda that takes into account constraints and avoids conflicts.   In practice, tasks come from three categories: 

\begin{itemize}[itemsep=1pt,topsep=2pt]
\item \textbf{Pre-scheduled} are events already scheduled for the day on one's calendar (e.g., work meetings and appointments).  
\item \textbf{Habits} are things that the user performs daily, such as eating meals and getting exercise.
\item \textbf{ToDos} are tasks that the user wants to accomplish for the day, such as reading a book, mowing the lawn, etc.
\end{itemize}
Ideally, the events that are fed to the planner for a given day should be automatically populated from the tools that the person is already using, such as their calendar or a  todo list tool. The habits can be inferred from the user's past activities when those are available. 

Formally, an input event to the planner, $e$, is a tuple of the form $(d, L)$ where $d$ is the event 
description (e.g., ``breakfast'') and $L = \{l_1, \dots, l_m\}$ is the list of constraints that the planner needs to satisfy. The output of the planner is an agenda 
consisting of a sequence of $m$ calendar events $\{c_1, \dots, c_m\}$
where each $c_i$ is a triplet $(e, \mathsf{start}, \mathsf{end})$
where $e \in E$ and $\mathsf{start} / \mathsf{end}$ are the  start/end
time of the event $e$.

We consider three types of  constraints (see Table~\ref{tab:constraints} for examples):
\begin{itemize}
    \item \textbf{Absolute temporal constraints}: These constrain the absolute start and end times of events and take the following form: $\{\mathsf{``starts\ at"}, \mathsf{``ends\ at"}, \mathsf{``before"}, \mathsf{``after"}\}$ + time (e.g., ``9am'').
    \item \textbf{Duration constraints}: These constrain only the duration of an event and are of the form: $\{``<", ``>"\}$ + duration (e.g., ``1 hr'').
    \item \textbf{Relative temporal constraints}: These constrain the relative ordering of pairs of events and take the form: $\{\mathsf{``before"}, \mathsf{``after"}\}$ + event.
\end{itemize}

\begin{table}[t]

\caption{\small Types of constraints on events for day planning. }
\label{tab:constraints}
\small
\centering
\begin{tabular}{cc} \toprule
Constraint type    & Examples\\ \midrule
Absolute temporal & starts at 9am, ends at 15:00, \\ 
 & before 5:00pm, after 2:00pm \\ \hline 
Duration & \textless{}90 minutes, \textgreater{}1 hour \\ \hline
Relative temporal & after lunch \\ \bottomrule
\end{tabular}
\vspace{-3mm}
\end{table}


Next, we describe each of the planners we built and their hybrid variants. 

\subsection{LLM-based planner (\sname)}
\label{sec:llm-planner}


The key idea underlying  \sname\ is to leverage LLM's commonsense reasoning capabilities and its ability to create structured results (in our case, plans). 
In \sname, we leverage LLMs for day planning via in-context learning~\cite{DBLP:conf/acl/LoganBWP0022}.
Specifically, we construct a prompt designed to enable the LLM to generate
correct plans. The prompt consists of 3 components: (1) a task description describing
the high-level goal of the task, (2) the list of input events and constraints formatted in text,
and (3) a 1-shot example of a plan. The details of the prompt are shown in Figure \ref{fig:gptprompt} in the
Appendix. Note that additional processing steps are necessary to correctly
extract the event name and start/end time from the LLM's output.

There are two points worth noting here. First, a traditional symbolic planner requires all the constraints as input. One of the benefits of using LLMs for planning is that the user does not need to specify all the constraints and some can be inferred from the commonsense knowledge in the LLM (e.g., appropriate times and lengths for meals or other activities).  Second, we expect that the interaction with the planner may involve several iterations, where the user refines their needs as they go along. Here too, the \sname\ will offer the benefit that it can understand user needs that are stated more vaguely in natural language, such as
``I want frequent breaks in the afternoon''. 


\subsection{Symbolic planner (\og)}
\label{sec:symbolic-planner}

We contrast  the LLM-based approach to planning with a symbolic planner, \og.  Given the set of events $E$ and constraints $L_e$ for each $e \in E$, \og computes the start and end times ($\mathsf{start}_e$ and $\mathsf{end}_e$) of the events. Since there can be multiple plans that satisfy all the constraints, \og chooses between them by solving an optimization problem that minimizes the total length of the schedule. Specifically, \og\ finds $\min  \{ \max_{e \in E} \mathsf{end}_e \}$  s.t., $l_{e,i}$ holds  $\forall l_{e,i} \in L_e, \forall e \in E$.



\begin{figure}[t]
    \centering
    \includegraphics[width=0.4\textwidth]{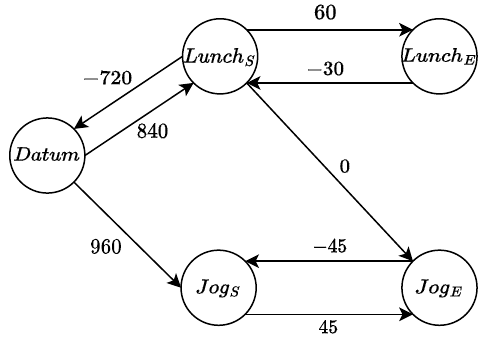}
    \caption{\small An example simple temporal network (STN) with start and end nodes for two events ($\mathsf{Lunch}$ and $\mathsf{Jog}$) and the $\mathsf{datum}$ ($t=0$). The edges represent temporal constraints, e.g., (a) the edges between $Lunch_S$ and $Lunch_E$ represent that its duration must be between 30 to 60 minutes, (b) $\mathsf{Jog}$ must start before 960 minutes from the datum i.e. before 4pm, and (c) $Jog_E$ must happen before $Lunch_S$.}
    \label{fig:stp}
\vspace{-3mm}
\end{figure}

Users don't always provide a consistent set of constraints. Hence, \og\ begins by employing a constraint checking and relaxation step. It begins
by encoding the events as nodes and temporal constraints as edges into the distance graph representation of a simple temporal network (STN)~\cite{dechter1991temporal} (see Figure~\ref{fig:stp}). The STN  allows us to obtain upper and lower bound times for scheduling each node by running shortest path algorithms on it. If we do not get valid bounds for an event, we use a procedure similar to \citet{oei2009resolving} to detect conflicting edges and either relax them or drop them completely if they cannot be relaxed.

Having obtained valid time bounds for scheduling each node and having asserted feasibility, we plan using a backtracking search which implicitly enforces a non-overlap between events in time. The search algorithm uses: (a) our distance graph representation for constraint propagation, and (b) slack-based heuristics to choose events which have a tighter remaining scheduling window earlier during the search. Upon completion of the search, we return the best full (or partial) solution found till then.
For more details, see Appendix~\ref{sec:app_og_plan}.

\subsection{Hybrid planners}

\smallskip
\noindent
\textbf{\sname + self-reflection: }
In our initial experiments with \sname, we found that it can follow instructions and
generate plans that are mostly correct (e.g., constraints are satisfied, events are non-overlapping).
It can also generate sensible plans even when the input constraints are under-specified.
For example, without specifying the start/end time of the ``breakfast'' event, the LLM
still knows to schedule the event in the morning. However, we also found that the planner may
miss some trivial but important constraints. For example, for pre-scheduled events with 
fixed start/end time, the model may still incorrectly place the event hours off the scheduled time slot.
To address this issue, we  apply additional feedback and reasoning steps to allow the LLM to self-correct
its output~\cite{shinn2023reflexion}.

 Specifically, after receiving an output plan from the LLM, we run a set of constraint checkers that identify a few kinds of violations: (1) unsatisfied input constraints, (2) missing events, and (3) overlapping events.
We then verbalize the list of violations into textual feedback such as ``the duration of A is too 
long/short'', ``A should not overlap with B'', etc., and feed them as input to the LLM to
re-generate the plan. We  repeat this process for a fixed number of iterations or until
there is no violation detected.
Among all the iterations, we select the plan with the minimal number of violations as the final
output. Details of the self-reflection prompt are in Figure \ref{fig:self_reflection}
of the appendix.

\smallskip
\noindent
\textbf{\llmog:} We extended \og\ to infer commonsense constraints about the durations and time of events from an LLM so it can compete with \sname's built-in ability to do so. Unless the timing of an event is completely specified, \llmog\ first queries the LLM to inquire about additional possible constraints (see the prompt in Appendix~\ref{sec:app_gptog_plan}). These additional constraints, as long as they don't conflict with the user-provided constraints, are then appended to the user-specified constraints for each event and are passed to \og as before.

%% file: conclusion.tex
\section{Conclusion} \label{sec:conclusion}

We considered  using LLMs to create a day plan. LLMs offer a promising technology for day planning and other human-centered tasks because user needs are specified with (sometimes, vague) natural language commands and may require multiple turns with the planner.   Our main result is to show that LLM-based planning is competitive  with  symbolic planning when constraints are explicit, but that the LLM is also able to infer missing constraints from its commonsense knowledge.

Our experiment with real users showed that some  constraints can be quite complex, and more research is needed to help the LLM to handle those effectively. For example, a request to ``break up a meeting into two parts'' may require a few steps to incorporate into the plan. Chain-of-thought reasoning may be a promising direction to explore. Another important research direction is to incorporate the user's context, such as their life situation, preferences, habits, and how they have handled similar constraints on previous days.
